# A Novel Multi-task Deep Learning Model for Skin Lesion Segmentation and Classification

*Xulei Yang[1], Zeng Zen[2], Si Yong Yeo[1], Colin Tan[3], Hong Liang Tey[3], Yi Su[1]*

1. Institute of High Performance Computing, A*STAR, Singapore
2. Institute for Infocomm Research, A*STAR, Singapore
3. National Skin Center, National Healthcare Group, Singapore

## ABSTRACT

In this study, a multi-task deep neural network is proposed for skin lesion analysis. The proposed multi-task learning model solves different tasks (e.g., lesion segmentation and two independent binary lesion classifications) at the same time by exploiting commonalities and differences across tasks. This results in improved learning efficiency and potential prediction accuracy for the task-specific models, when compared to training the individual models separately.

The proposed multi-task deep learning model is trained and evaluated on the dermoscopic image sets from the International Skin Imaging Collaboration (ISIC) 2017 Challenge "*Skin Lesion Analysis towards Melanoma Detection*", which consists of 2000 training samples and 150 evaluation samples. The experimental results show that the proposed multi-task deep learning model achieves promising performances on skin lesion segmentation and classification. The average value of Jaccard index for lesion segmentation is 0.724, while the average values of area under the receiver operating characteristic curve (AUC) on two individual lesion classifications are 0.880 and 0.972, respectively.

## 1. INTRODUCTION

Clinical research has shown that skin cancer is a major public health problem. It is estimated that one in five Americans may be affected by skin cancer [1]. Melanoma is one of the most common type and dangerous skin cancer. Early detection and accurate diagnosis of skin lesion is crucial, in particular for melanoma.

Dermoscopic evaluation of melanocytic lesions (or moles) to detect melanoma is the current standard in clinical practice. Dermatologists specialised in managing skin disease see patients for concerns about new or changing moles and lesions. Specific features are to be detected under the dermoscope and the features are tabulated using validated algorithms to help determine the risk of melanoma [2, 3] by the dermatologists. However, the multitude of dermoscopic features and algorithms are complex, confusing and tedious to detect [4]. As such, many dermatologists do not utilize the dermoscopic tools properly or accurately, which may cause compromised clinical care.

Different imaging techniques, such as multispectral imaging and confocal microscopy, have been implemented to address the issues encountered in the detection of melanomas. However, such imaging devices are costly and bulky, and the dermatologist has to be trained in these imaging modalities. It is shown that dermoscopic examination by trained and experienced dermatologists yields better sensitivity and specificity [5] in skin lesion diagnosis. Therefore, an automatic technique for robust analysis of dermoscopic dataset can be advantageous to clinicians.

Advanced dermoscopic algorithms [6, 7], such as "chaos and clues," "3-point checklist," "ABCD rule," "Menzies method," "7-point checklist," and "CASH" had been developed to facilitate a novice's ability to distinguish melanomas from benign with high diagnostic accuracy. Among these clinical evaluation algorithms, studies have shown that pattern analysis yields better diagnostic performance over other approaches [8]. The performance of pattern analysis for melanoma detection, however, greatly depends on the choice of meaningful descriptive features derived from the dermoscopic images. Identification of such features requires domain-specific expert knowledge and may fail for complex image segmentation and classification problems.

Deep neural networks techniques [9], in particular, the deep convolutional neural networks (DCNN) [10], have dramatically improved the state-of-the-art in object categorization and object detection. The DCNN has also been widely used on biomedical dataset, such as for skin lesion analysis [11, 12]. Different features detected at the different convolutional layers allow the network to handle large variations in the dataset. It enables the feature detection to be handled automatically, thus ameliorating the difficulties of feature detection inherent in conventional pattern analysis techniques.

In this article, a novel multi-task DCNN model will be used for the analysis of the skin lesion. Different from conventional DCNN, in the multi-task learning architecture, the input dermoscopic image can be associated with multiple labels that describe different characteristics of the lesion. This multi-task technique can then be used in the segmentation of lesion and classifications of lesion categories at the same time with improved learning efficiency and prediction accuracy. This equips the dermatologist with a robust tool for analyzing dermoscopic dataset.

## 2. METHODOLODY

In this section, the multi-task DCNN model for simultaneous segmentation and classification of skin lesion will be described.

### 2.1. DCNN

The DCNN is a set of neural networks with many different convolutional layers that can be trained to extract features from images. During the training phase, the DCNN will minimize the difference between its current model and the labelled components. In the process, features of different scales are also extracted at the different convolutional layers. This will allow for robust detection and categorization of structural components in the evaluation of a new input image. The DCNN technique can therefore be used in the categorization of skin lesion in a dermoscopic dataset.

The DCNN technique is, however, not a widely used tool by dermatologists in skin lesion analysis. This is due to challenges posed by artefacts, aberrations and noise in the dermoscopic dataset, which prevents robust detection of melanoma. In addition, there is large variation in the dermoscopic dataset, in which skin lesion may contain similar structural features and color. For example, the seborrheic keratosis (SK) – which is a benign disease – can be difficult to distinguish from melanoma. Therefore, a large training dataset has to be used in the training of conventional DCNN model, and this creates roadblocks for practical application of the DCNN technique.

### 2.2. The multi-task DCNN model

The multi-task DCNN technique allows different components to share the detected features among different attribute categories. These components will generate feature representations specific to the attribute categories, and a multi-task training on the features can be used to infer the attributes. This will allow robust detection of the skin lesion from a dermoscopic dataset.

The segmentation of the lesion from dermoscopic images is an important aspect of melanoma detection, as some of the features which are used by clinicians in dermoscopy algorithms are based on the shape of the skin lesion. The segmentation of the lesion is therefore incorporated in the multi-task network model for skin lesion analysis.

The diagram of the proposed multi-task DCNN technique is depicted in Fig.1. The network consists of different components that will be trained using an annotated dataset by a clinician. The annotation consists of (a) the categorization of the skin lesion, and (b) the delineation of the lesion shape. In particular, the network consists of three components for the robust analysis of the skin dataset:

(1) Component-1 is used for the segmentation of the boundary of the skin lesion in the dermoscopic data.
(2) Component-2 is used for the categorization of dermoscopic data into (a) melanoma and (b) nevus and seborrheic keratosis.
(3) Component-3 uses the detected features for categorizing the dermoscopic images into (a) seborrheic keratosis and (b) nevus and melanoma.

The detected features learned by the DCNN can be used by the different components in the multi-task inference, i.e., the segmentation of the lesion and the detection of melanoma in the skin dataset. The output of the multi-task network includes (a) the binary mask of the skin lesion, (b) the probability of detected skin lesion belonging to melanoma, and (c) the probability of detected skin lesion belonging to seborrheic keratosis.

The proposed multi-task DCNN model is implemented based on the layout architectures of GoogleNet [13] and U-Net [14]. The layout structure of the multi-task DCNN model is depicted in Fig. 2. We will share the source code of the multi-task deep learning model via GitHub in the upcoming version of this study.

## 3. EXPEERIMENT AND DISCUSSION

**Datasets and Tasks:**
The training and evaluation dermoscopic datasets are from the ISIC 2017 Challenge "Skin Lesion Analysis towards Melanoma Detection". The background of this challenge can be found in last year's event [15]. There are 2000 samples in the training dataset and 150 samples in the evaluation dataset. The goal of the challenge is to help participants develop image analysis tools to enable automatic diagnosis of melanoma from dermoscopic images. There are three tasks in this year's competition:

(1) Task-1: Skin lesion segmentation,
(2) Task-2: Detection and localization of visual dermoscopic features, and
(3) Task-3: Disease Classification.

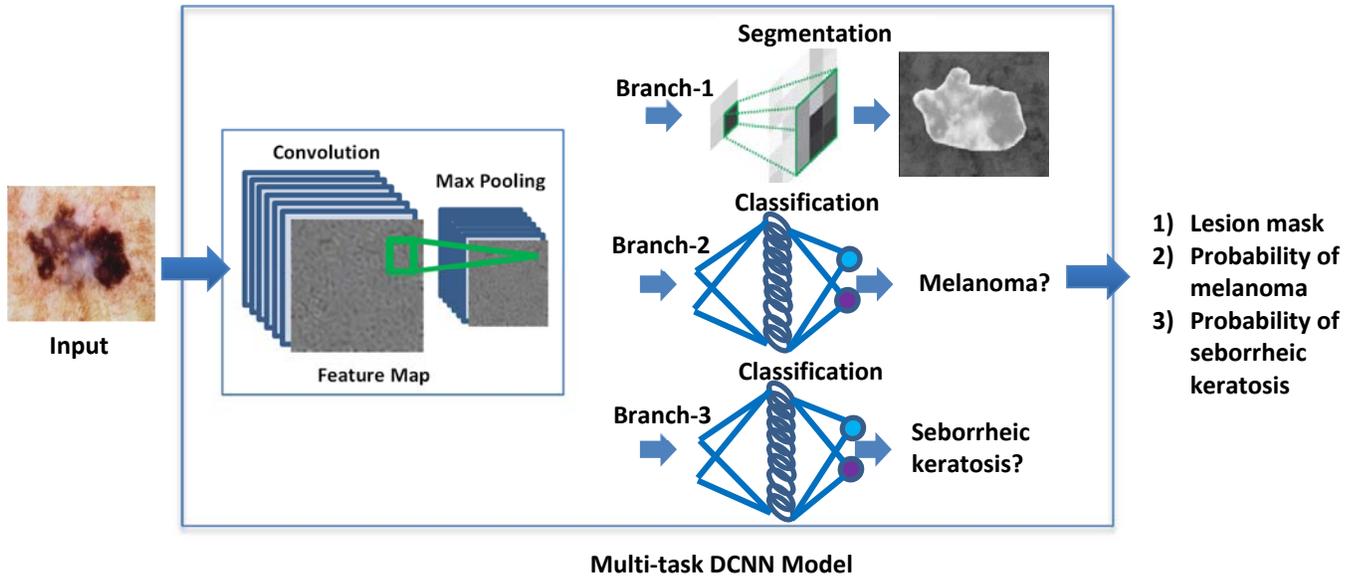

**Fig. 1** Diagram of proposed multi-task DCNN model for skin lesion analysis.

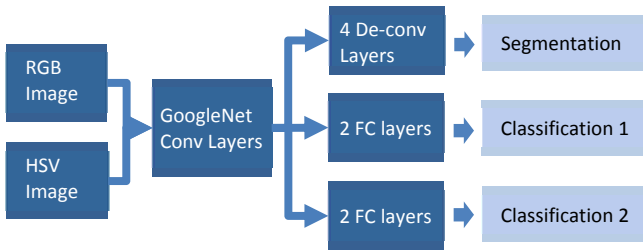

**Fig. 2** Layout structure of proposed multi-task model.

Task-3 includes two individual classification tasks: The first is to distinguish between (a) melanoma and (b) nevus and seborrheic keratosis; The second is to distinguish between (a) seborrheic keratosis and (b) nevus and melanoma. In this work, we focus on Task-1 and Task-3 only. We leave Task-2 for future research work.

During training, the samples are augmented by rotation and flipping, and pre-processed by zero mean unit standard deviation normalization. During testing, the samples are also pre-processed by zero mean unit standard deviation normalization.

**Evaluation Metrics:**
The segmentation results are evaluated using the Jaccard Index, also known as Intersection-over-Union (IoU). The IoU is a measure of overlap between the area of the automatically segmented region and that of the manually segmented region. The value of IoU ranges from 0 to 1, with a higher value implying better match between the two regions.

The classification results are evaluated by the area under the receiver operating characteristic curve (AUC). The AUC is a measure of how well a parameter can distinguish between two diagnostic groups (diseased/normal).

**Experimental Results:**
The multi-task network is compared against conventional DCNN, e.g., GoogleNet for lesion disease classification and UNet for lesion segmentation. In this study, a 5-fold cross-validation is used to train and evaluate all the deep learning models. The trained deep learning models are tested on 150 evaluation samples and the results are submitted to the challenge website for numerical assessment based on the aforementioned evaluation metrics.

Fig. 3 illustrates the output of the proposed multi-task model on evaluation sample "ISIC_0012484" (Fig. 3a). The model simultaneously produces three results including segmentation (Fig. 3b), probability of melanoma (Fig. 3c), and probability of seborrheic keratosis (Fig. 3d).

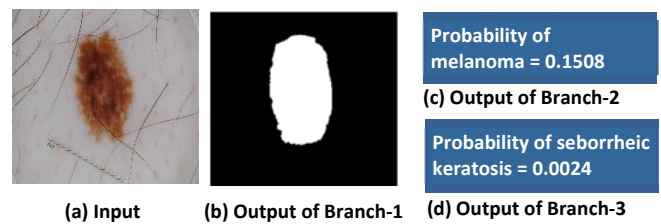

**Fig. 3** Output of proposed multi-task model on skin lesion image ISIC_0012684.

Table 1 summarizes the average values of Jaccard Index (IoU) achieved by the proposed multi-task model and the well-known U-Net model [14]. Table 2 summarizes the average values of AUC achieved by the proposed multi-task model and the well-known GoogleNet model [13].

Table.1 Performance comparison between proposed multi-task model and U-Net on ISIC 2017 Challenge Task-1: lesion segmentation

| Method | The proposed model | U-Net [14] |
|---|---|---|
| Jaccard Index | 0.724 | 0.717 |

Table 2 Performance comparison between proposed multi-task model and GoogleNet on ISIC 2017 Challenge Task-3: disease classification

| Method | The proposed model | GoogleNet [13] |
|---|---|---|
| AUC value | 0.926 | 0.903 |

**Discussions:**
From the comparisons in Table 1 and Table 2, it was observed that the proposed multi-task deep learning model achieves slightly better segmentation and classification performance as compared to individual U-Net and GoogleNet models, respectively. The segmentation appears to boost the classification performance quite a bit, but the converse seems not to have significant effect. Nevertheless, it has to be highlighted that the proposed model integrates multiple tasks into one single model to achieve much higher training and testing efficiency in terms of computation cost.

## 4. CONCLUSION

In this article, a multi-task DCNN technique is described for the analysis of skin lesion. The network consists of different components for the robust analysis of the skin lesion. The multi-task technique allows the different components to share the detected features among the different attribute categories. The multi-task network technique is more robust and efficient as compared to the conventional DCNN technique. The incorporation of the different annotations of the skin dataset allows the network to detect the features which can describe the different attribute of the skin dataset. This allows a robust detection of the lesion in the skin dataset.